\documentclass{article}

    \PassOptionsToPackage{numbers, compress}{natbib}
 \usepackage[preprint]{neurips_2026}

\usepackage[utf8]{inputenc} 
\usepackage[T1]{fontenc}    
\usepackage{hyperref}       
\usepackage{url}            
\usepackage{booktabs}       
\usepackage{amsfonts}       
\usepackage{nicefrac}       
\usepackage{microtype}      
\usepackage{xcolor}         
\usepackage{wrapfig}
\usepackage{siunitx}
\usepackage{graphicx}
\usepackage{adjustbox}
\usepackage{algorithm}
\usepackage{algorithmic}
\usepackage{amsmath}
\title{BubbleSH: A Dataset of Rising Bubbles with Deformable Interfaces}

\definecolor{darkblue}{rgb}{0.0, 0.0, 0.55}
\hypersetup{
    colorlinks=true,
    citecolor=darkblue, 
    linkcolor=darkblue, 
    urlcolor=darkblue   
}

\author{%
Rachna Ramesh$^{1}$ \quad
Kiet Bennema ten Brinke$^{1}$ \quad
Douwe Orij$^{2}$ \quad
Ivo Roghair$^{2}$ \quad \\
\textbf{Vlado Menkovski}$^{1}$ \\[2mm]
{\small
$^{1}$ Department of Mathematics and Computer Science,
Eindhoven University of Technology} \\
\small{
$^{2}$ Department of Chemical Engineering and Chemistry,
Eindhoven University of Technology} \\[1mm]
{\small
\texttt{\{r.ramesh1, k.bennema.ten.brinke, d.r.orij,  i.roghair, v.menkovski\}@tue.nl} 
}
}

\begin{document}

\maketitle

\begin{abstract}
  Bubbly flows exhibit complex multiscale dynamics, with deformable bubbles interacting through the surrounding liquid and giving rise to strongly coupled kinematic and morphological behavior. We present BubbleSH, a bubbly flows dataset consisting of transient, three-dimensional bubble-swarm dynamics obtained from high-fidelity direct numerical simulations of bubbles rising in a periodic domain. The dataset provides time-resolved bubble trajectories, velocities, and shape evolution, with bubble morphology compactly represented using spherical harmonics. Designed to be lightweight yet physically expressive, the dataset enables data-driven modeling of bubbly flow simulators where shape deformation and bubble–bubble interactions play a central role. We characterize the dataset with bubble kinematics, morphology, and interaction patterns, and introduce evaluation metrics for both trajectory and shape prediction. The sensitivity of bubble-swarm dynamics to local perturbations makes BubbleSH particularly well suited to generative models that learn distributions over possible future trajectories. We evaluate a permutationally and translationally equivariant probabilistic emulator on BubbleSH given the proposed metrics. Therefore, we establish a compact, high-fidelity dataset and a benchmark for developing and evaluating data-driven models of deformable, chaotic multiphase systems. The dataset is publicly available on Zenodo under DOI: \href{https://doi.org/10.5281/zenodo.21229301}{https://doi.org/10.5281/zenodo.21229301}.
\end{abstract}

\section{Introduction}
\begin{wrapfigure}[28]{R}{0.41\textwidth}
    \includegraphics[width=1.00\linewidth]{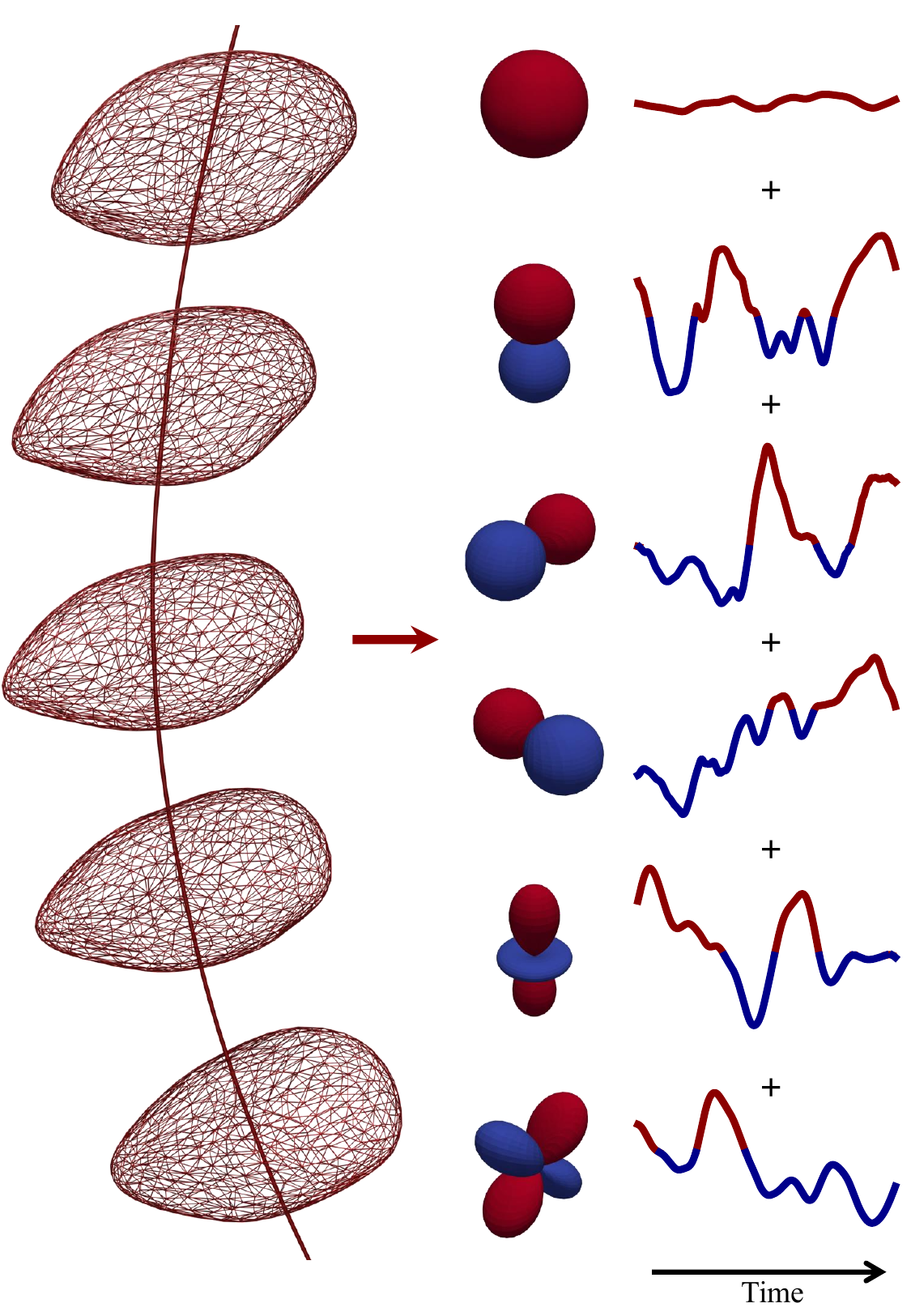}
    \caption{BubbleSH contains trajectories of bubble centroids (left) and spherical harmonics coefficients (right), calculated from simulated interface meshes (left)}
    \label{front}
\end{wrapfigure}
Bubbly flows, which consist of gas bubbles that rise through a liquid, occur in a wide range of natural and industrial systems.  They play a central role in chemical and biochemical reactors, metallurgical processes, wastewater treatment, and many other applications. In bubble swarms, rising bubbles influence each other’s trajectories, generate turbulence, induce macroscopic circulation patterns in the surrounding liquid and affect mass, momentum and energy exchange between phases. A detailed understanding of bubble motion, deformation, and interactions within swarms is essential to design efficient reactors, improve scale‑up methodologies, and advance predictive simulation tools.

Obtaining such insights experimentally remains extremely challenging due to optical occlusion and dense multiphase interactions, especially at higher gas fractions. For this reason, significant emphasis has been placed on numerical approaches, in particular Direct Numerical Simulations (DNS), with modern interface‑resolving methods such as Volume‑of‑Fluid (VoF) \cite{Hirt1981,Youngs1982,Rider1998,SintAnnaland2005} and Front‑Tracking (FT) \cite{Unverdi1992,SintAnnaland2006,Roghair2016}. The FT method explicitly tracks deformable bubble interfaces and avoids the (unphysical) numerical coalescence present in VoF schemes. FT has been demonstrated to reproduce complex bubble interaction behavior, for instance wobbling bubbles and path instabilities \cite{Dijkhuizen2010a,Dijkhuizen2010b}, bubble–bubble interactions and swarm statistics \cite{Roghair2013aiche,Bunner2002a,Bunner2002b,Tagawa2013}, turbulence generation \cite{Roghair2011ijmf}, and providing physically consistent drag forces \cite{Roghair2011ces} and mass transfer rates \cite{Darmana2006,Roghair2016} over a wide range of gas fractions. 


While Direct Numerical Simulations (DNS) provide an exceptional level of detail, their applicability is limited to small scales due to computational cost. Larger-scale simulations therefore rely on Euler–Lagrange (EL) models, which describe the liquid phase through detailed closures but approximate the bubble phase using time-averaged correlations for drag, lift and virtual mass derived from DNS or experimental data. This averaging collapses rich information on bubble-to-bubble variability, history effects, transient deformation, and local interactions into effective coefficients, eliminating variance in rise velocity, morphology, and trajectories.


This loss of information motivates data-driven approaches that aim to retain the fidelity of DNS while enabling scalable modeling. However, a critical bottleneck in this direction is the lack of high-quality, openly available datasets that capture transient, three-dimensional bubble dynamics at the level of individual bubbles within swarms. Although DNS produces such data, it is rarely disseminated in reusable form due to its size and complexity. To address this gap, we introduce a compact, high-fidelity dataset, BubbleSH, of deformable bubbles rising in swarms, derived from interface-resolving FT DNS. The data set provides a structured, time-resolved representation of bubble motion and deformation, while remaining lightweight enough for downstream analysis and learning. 

As such, BubbleSH provides a unique opportunity for developing machine learning models for dynamical systems that combine n-body interaction with deformable surfaces. Interacting-particle benchmarks are commonly used for the development and evaluation of geometric deep learning models. These datasets typically consist of multiple objects whose dynamics are coupled through pairwise and collective interactions \cite{kipf2018, egnnSatorras}. In contrast,  BubbleSH includes deformable objects that evolve in a turbulent flow. Here, the learning problem includes both coupled trajectories and time-varying surface geometry. Therefore, BubbleSH combines challenges from two established classes of benchmarks: first, point-particle datasets in which objects are rigid and do not carry shape information, and second, continuum simulation benchmarks, where deformation is represented by discretizing the domain into particles or a mesh (e.g., GNS \cite{GNSSanchez_Gonzalez}, MeshGraphNets \cite{MeshGraphNets} and LagrangeBench \cite{LagrangeBench}). BubbleSH is further unique, as the spherical nature of the bubbles lends for effective non-discretized representation of the shape of the surfaces as a truncated spherical harmonic expansion. Finally, the sensitive nature of bubbly flows due to the turbulent properties of the fluid with which they interact motivates the use of probabilistic modeling \cite{hendriks2026}. These properties make BubbleSH a natural benchmark for data-driven emulators. 

In this paper, we introduce a new dataset of deformable bubbles that rise in swarms, derived from high‑resolution FT DNS of air–water systems. The trajectories capture the time resolved evolution of bubble swarms under periodic boundary conditions, where each bubble is parameterized by its positions, velocity, and the coefficients of the spherical harmonic orbital functions. Alongside the dataset, we provide analysis and benchmarking tools for downstream model development. For each simulation configuration, we characterize the resulting dynamics through statistics of bubble motion, shape evolution, and inter-bubble interactions. These statistics are then used to define evaluation metrics for learned emulators, combining trajectory and shape-level errors with distributional comparisons that reflect the stochastic nature of bubble-swarm dynamics. Additionally, as an initial benchmark, we have trained a probabilistic generative forecasting model, positioning BubbleSH as a benchmark for probabilistic geometric modeling of deformable multiphase dynamics.  

\section{Related work}
 Existing bubble datasets are concentrated primarily in the image domain. Early work such as BubGAN introduced large collections of synthetic labeled bubbly-flow images for benchmarking detection and segmentation methods \cite{BubGAN}, while subsequent studies used experimental or semi-synthetic image corpora to develop deep-learning pipelines for bubble detection, mask extraction, and tracking in two-phase flows and boiling systems \cite{Soibam, kimDeepLearningBasedAutomated2021}. 
More recent work has moved toward 3D bubble reconstruction and tracking: Hessenkemper et al. introduced a semi-artificial dataset for 3D detection and tracking of deformable bubbles in swarms, and Yu et al. released 3DBubbles, an experimental dataset containing reconstructed 3D bubble structures together with corresponding 2D projections \cite{Hessenkemper3D, yu3DBubblesExperimentalDataset2026}. These datasets are oriented primarily toward image analysis, reconstruction, or tracking. 


A second line of related work concerns the use of spherical harmonics for compact representation of 3D surfaces. They have been used for surface parametrization and geometric modeling of closed shapes \cite{brechbuhlerRepresentation3DShape1992a, nortjeSphericalHarmonicsSurface2015a, zhou3DSurfaceFiltering2004, shenLargeScaleModelingParametric2006a}. In bubbly-flow research, they have been applied to characterize oscillation modes and interface distortions \cite{aganinDynamicsGasBubbles2022, brennShapeOscillationsPath2006, cattaneoShapeModesJet2025, inserraMathematicalModellingAcoustic2025, lalanneNonlinearShapeOscillations2015}. These works demonstrate that SH coefficients provide a compact and physically meaningful description of bubble morphology, particularly through lower-order modes associated with large-scale deformations. However, prior work has primarily used SH as a tool for analysis or static reconstruction of individual shapes. Their use as the underlying state representation in time-resolved datasets remains limited. Recent efforts such as 3DBubbles \cite{yu3DBubblesExperimentalDataset2026} move toward data-driven applications, but remain focused on experimental reconstruction rather than the dynamics of interacting bubble swarms derived from DNS.

\section{Data Generation}

The simulations underlying the present dataset are performed with a three‑dimensional FT method for incompressible two‑phase flow, which explicitly resolves deformable gas–liquid interfaces. 

\subsection{Front-Tracking Simulations}
The flow field is governed by the incompressible Navier–Stokes equations written in a one‑fluid formulation, in which both phases are described on a single Eulerian grid and the interface effects enter as localized source terms. The governing equations read

\begin{equation}
\rho(\mathbf{x})\left(\frac{\partial \mathbf{u}}{\partial t} + \mathbf{u}\cdot\nabla \mathbf{u}\right)
= -\nabla p + \nabla\cdot\left[\mu(\mathbf{x})\left(\nabla \mathbf{u} + \nabla \mathbf{u}^T\right)\right]
+ \rho(\mathbf{x})\mathbf{g} + \mathbf{f}_\sigma, \ \ \ \nabla\cdot\mathbf{u} = 0
\end{equation}






where $\mathbf{u}$ is the velocity field, $p$ the pressure, $\rho(\mathbf{x})$ and $\mu(\mathbf{x})$ the spatially varying density and viscosity, $\mathbf{g}$ gravity, and $\mathbf{f}_\sigma$ the surface tension force acting at the gas–liquid interface. The equations are discretized on a staggered Cartesian grid using a finite‑volume approach. A velocity projection method with iterative pressure correction is employed to enforce incompressibility. All linear systems arising from the discretization are solved with an incomplete‑Cholesky conjugate‑gradient (ICCG) solver, providing robustness and efficiency for the large three‑dimensional problems considered.

The gas–liquid interface (front) is explicitly represented (tracked) by a moving triangular surface mesh, whose nodes are advected with the local fluid velocity. Velocities at the nodes are obtained by piecewise spline‑based interpolation from the Eulerian grid, and time integration of the interface position is performed using a fourth‑order Runge–Kutta (RK4) scheme. This explicit tracking yields an accurate description of interface deformation dynamics and curvature, which is essential for reliable surface tension calculations.

\begin{wrapfigure}[30]{R}{0.38\textwidth}
\vspace{-15pt}
\adjustbox{trim=0cm 0.4cm 0cm 0.0cm}{
    \includegraphics[width=\linewidth]{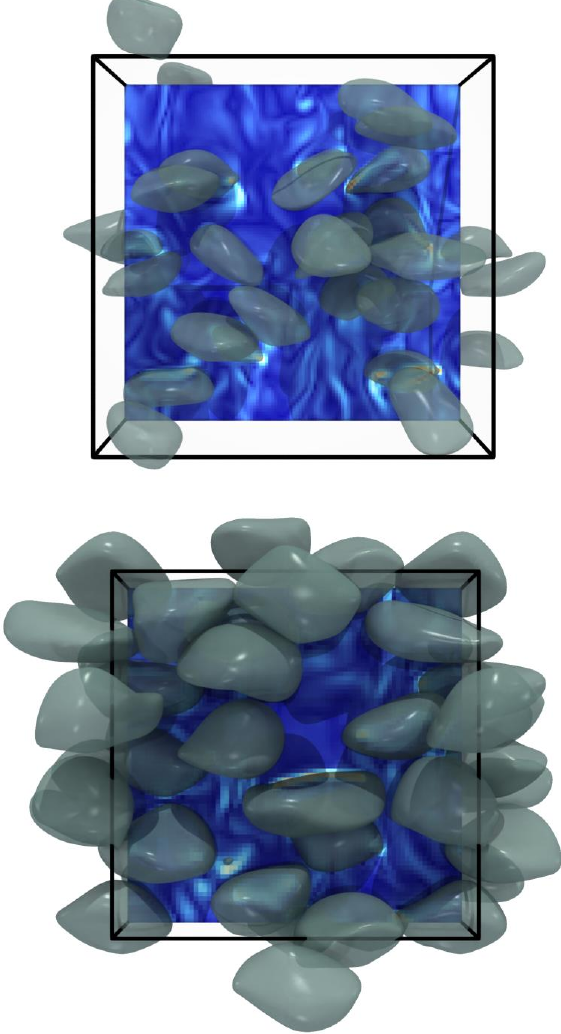}
}
\caption{Snapshots of simulations of \(d=\qty{6}{\milli\meter}\) bubbles with void fractions of 0.10 (top) and 0.30 (bottom). The background slice indicates the vorticity of the liquid phase.}
\label{sim}
\end{wrapfigure}

To maintain a high‑quality surface representation during long transient simulations, the interface is continuously remeshed. The primary operation is smoothing, which redistributes marker points so that triangles remain nearly equilateral and the accuracy of curvature and surface tension forces is preserved. Mesh quality is further controlled through local topological operations, such as node addition, node removal, and edge swapping, based on local edge length and curvature criteria. Any local change that affects the enclosed volume is immediately corrected by volume‑preserving node repositioning, thereby ensuring both local and global conservation of bubble volume throughout the simulation.

The FT code simulates multiple bubbles in a single, fully periodic domain, effectively representing an infinite bubble swarm without wall effects. As bubbles approach, the intervening liquid film is naturally resolved and drains dynamically under hydrodynamic forces. Unlike front‑capturing methods such as Volume‑of‑Fluid or Level‑Set techniques \cite{SintAnnaland2005}, the FT method does not allow automatic interface merging; coalescence is therefore prevented by design, enabling bubble-swarm interactions to be studied without artificial merging. If numerical operations cause interfaces to intersect, these events are detected explicitly, and corrected by locally conservative node repositioning, again preserving bubble volume.

Surface tension forces are computed directly from the triangular mesh geometry and mapped onto the Eulerian grid as localized force densities. The exact interface position also allows an analytical determination of the local phase fraction in each computational cell, from which the spatially varying density and viscosity fields are constructed using appropriate averaging in mixture cells. This tight coupling between the Lagrangian interface description and the Eulerian flow solver is a defining feature of the FT approach and underpins its accuracy for deformable multiphase flows.

At prescribed output intervals, typically of order \qty{1E-4}{\second}, the code records detailed bubble dynamics, including the full surface mesh of each bubble, centroid positions, translational velocities, and additional geometric and hydrodynamic quantities. These high-fidelity outputs form the basis of the dataset presented in this work.

\subsection{Spherical Harmonics Representation}
    The raw FT output contains a detailed triangulated interface for every bubble at every recorded time step. A single bubble mesh typically contains \num{1.1E4} to \num{1.5E4} nodes (\num{2.2E4}--\num{3.0E4} triangular cells), distributed over the surface, depending on the amount of deformation.  Although this mesh-based representation is geometrically detailed, meshes are storage-intensive and inconvenient for downstream statistical analysis and tasks, which typically require compact, fixed-dimensional descriptors. In order to reduce the footprint with several orders of magnitude, the triangulated bubble surface meshes are projected onto a basis of spherical harmonics. 

Spherical harmonics provide a smooth and orthonormal basis on the sphere, and are well suited to bubble interfaces, which are typically near-spherical but may exhibit moderate deformations such as ellipsoidal or wobbling disk-like shapes. The bubble surface is described in a local spherical coordinate system centered at the bubble centroid. Let \((\theta,\phi)\) denote spherical coordinates, where \(\theta \in [0,2\pi]\) is the azimuthal angle and \(\phi \in [0,\pi]\) is the polar angle, and let \(r(\theta,\phi)\) be the radial distance from the centroid to the interface in direction \((\theta,\phi)\) \cite{yu3DBubblesExperimentalDataset2026}. The surface is approximated by a truncated spherical harmonic expansion,
\begin{equation}
r(\theta,\phi) \approx \sum_{\ell=0}^{L}\sum_{m=-\ell}^{\ell} c_{\ell m}\, Y_{\ell}^{m}(\theta,\phi),
\end{equation}
where \(Y_{\ell}^{m}\) are spherical harmonic basis functions given by Equation \ref{shbasis} and \(c_{\ell m}\) are the associated coefficients. 
\begin{equation}\label{shbasis}
Y_{\ell}^{m}(\theta,\phi)
=
\sqrt{\frac{(2\ell+1)(\ell-m)!}{4\pi(\ell+m)!}}
\,P_{\ell}^{m}(\cos\phi)\,e^{im\theta},
\end{equation}
where \(P_{\ell}^{m}\) denotes the associated Legendre polynomial \cite{yu3DBubblesExperimentalDataset2026}.
For a truncation order \(L\), the representation yields \((L+1)^2\) coefficients per bubble. Lower-order modes capture the dominant large-scale deformations of the interface, while higher-order modes encode progressively finer surface structure. The representation assumes that the bubble is star-like with respect to its centroid, so that each ray emanating from the centroid intersects the interface only once. This excludes extreme cases such as toroidal or highly concave interfaces, but is appropriate for the class of deformable bubbles considered in the present simulations.

The coefficients are computed independently for every bubble at every recorded time step. The vertices of each triangulated interface are first expressed relative to the bubble centroid and converted from Cartesian to spherical coordinates, producing sampled values of \(r(\theta,\phi)\) over the interface. A truncated spherical harmonic expansion is then fitted to these samples in a least-squares sense, and the resulting coefficient vector is stored as the shape descriptor of the bubble. This representation preserves the essential geometry of deformable bubble interfaces while reducing the dimensionality of the original mesh-based simulation.

\section{Dataset and Metrics}

\subsection{Data Content}

\begin{wraptable}[11]{r}{0.35\textwidth}
\vspace{-1em}
\caption{Physical properties used in the simulations.}
\vspace{0.6em}
\centering
\footnotesize
\renewcommand{\arraystretch}{1.02}
\setlength{\tabcolsep}{1pt}
\begin{tabular}{@{}lcc@{}}
\toprule
\textbf{Property} & \textbf{Symbol} & \textbf{Value} \\
\midrule
Gas density                & $\rho_g$ & \qty{1.25}{\kilogram\per\cubic\meter} \\
Gas viscosity              & $\mu_g$  & \qty{1.8e-5}{\pascal\second} \\
Liquid density             & $\rho_l$ & \qty{1000}{\kilogram\per\cubic\meter} \\
Liquid viscosity           & $\mu_l$  & \qty{1e-3}{\pascal\second} \\
Surface tension & $\sigma$ & \qty{0.073}{\newton\per\meter} \\
\bottomrule
\end{tabular}

\label{tab:physical_properties}
\vspace{-0.8\baselineskip}
\end{wraptable}

The dataset is organized as a collection of simulations of deformable bubble swarms in an air--water system rising under the influence of gravity. Each simulation corresponds to a fixed combination of bubble diameter \(d\) and gas volume fraction \(\varepsilon\), and records the time-resolved evolution of all bubbles in a cubic periodic domain. Periodic boundary conditions are applied in all three spatial directions, so bubbles exiting one side of the domain re-enter through the opposite side. See Table~\ref{tab:physical_properties} for more simulation details.

\begin{wraptable}[17]{r}{0.35\textwidth}
\vspace{-1em}
\centering
\caption{Parameter configurations covered by the BubbleSH dataset. Temporal resolution $\Delta t$ is listed for each $\varepsilon$ and  $d$ simulation parameter.}
\vspace{0.5em}
\footnotesize
\renewcommand{\arraystretch}{1}
\setlength{\tabcolsep}{3.5pt}
\begin{tabular}{@{}cccc@{}}
\toprule
\textbf{$\varepsilon$ (\%)} & \multicolumn{3}{c}{\textbf{$\Delta t$ (s)}} \\
\cmidrule(lr){2-4}
& \textbf{4 mm} & \textbf{5 mm} & \textbf{6 mm} \\
\midrule
 5  & $10^{-4}$ & $10^{-4}$ & $10^{-4}$ \\
10  & $10^{-4}$ & $10^{-4}$ & $10^{-4}$ \\
15  & $10^{-3}$ & $10^{-4}$ & $10^{-3}$ \\
20  & $10^{-4}$ & $10^{-4}$ & $10^{-4}$ \\
25  & $10^{-3}$ & $10^{-4}$ & $10^{-3}$ \\
30  & $10^{-4}$ & $10^{-4}$ & $10^{-4}$ \\
35  & $10^{-3}$ & $10^{-4}$ & $10^{-3}$ \\
40  & $10^{-4}$ & $10^{-4}$ & $10^{-4}$ \\
\bottomrule
\end{tabular}

\label{tab:dataset_configurations}
\end{wraptable}

The parameter space spans three bubble diameters, \num{4}, \num{5}, and \qty{6}{\milli\meter}, and eight gas volume fractions ranging from 5\% to 40\% in increments of 5\%, yielding 24 parameter configurations. For each configuration, the simulation tracks a swarm of \(N=32\) bubbles over time. The domain size $L_\mathrm{box}$ is determined by \(\varepsilon\) and bubble radius,
$L_{box}=(N \frac{4}{3} \pi r^3_{bub} \ / \ \varepsilon)^{\frac{1}{3}}$, where \(r_{\mathrm{bub}}=d/2\). $L_\mathrm{box}$ is in the range of 6 to \qty{41}{\milli\meter}.


At each time step \(t\), and for each bubble \(i\), the dataset contains the centroid position \(\mathbf{x}_i(t) \in \mathbb{R}^3\), velocity \(\mathbf{v}_i(t) \in \mathbb{R}^3\), and shape descriptor \(\mathbf{c}_i(t) \in \mathbb{R}^{225}\) \, represented by a fixed set of spherical harmonic coefficients. We use up to order $L=14$ of the spherical harmonics basis functions, yielding $ (14+1)^2=225$ coefficients per bubble. This truncation order was selected such that the mean absolute percentage error (MAPE) on the reconstructed surface area remains below $0.1\%$, while the reconstruction error decreases rapidly for increasing $L$ \cite{yu3DBubblesExperimentalDataset2026}. The FT simulator represents each bubble shape using $3.9\times10^3$ floats, compressing these to 225 spherical harmonic coefficients yields a compression ratio of $\sim 173\times$. These quantities describe both the translational motion of each bubble and its time-varying surface deformation.

Each trajectory covers between \qty{1.5}{\second} and \qty{2}{\second} of simulated time. The temporal resolution depends on the parameter configuration and is set to either \qty{0.1}{\milli\second} or \qty{1}{\milli\second}. The initial \qty{0.2}{\second} are discarded to eliminate startup transients, resulting in a usable duration of approximately \qtyrange{1.3}{1.8}{\second}. Table~\ref{tab:dataset_configurations} summarizes the main characteristics of the dataset and the parameter ranges considered. Visualizations are provided in Appendix \ref{app:viz}.

\begin{figure}
\adjustbox{trim=1.1cm 0.4cm 0cm 0.8cm}{
    \includegraphics[width=1.08\textwidth]{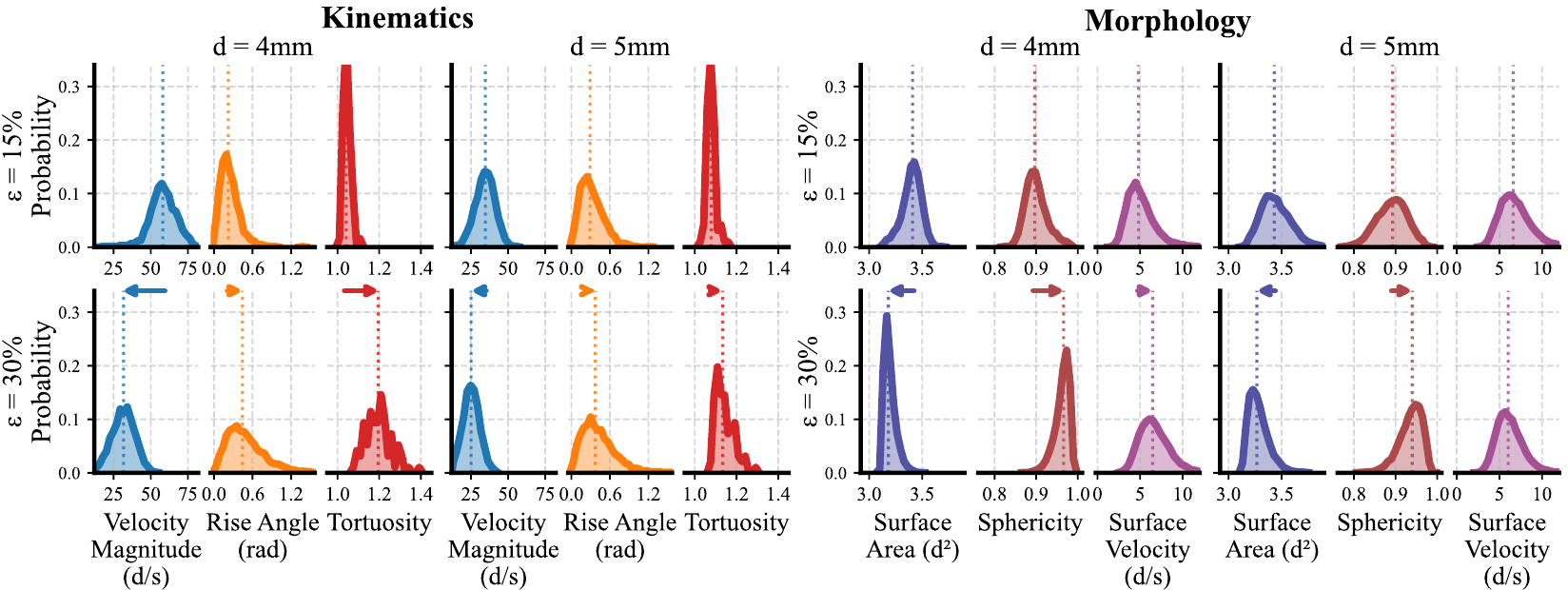}}
    \caption{Probability mass functions of kinematic and morphological quantities defined in Section \ref{statistics} for two different bubble diameters $d$ and gas volume fractions $\varepsilon$. Dotted line indicates the median, arrows indicate median shift, number of bins is 30. Unit of length is bubble diameter $d$.}
    \label{pmf1}
\end{figure}

\subsection{Dataset Statistics} \label{statistics}
To characterize the dynamics captured in BubbleSH, we analyze kinematic, morphological, and interaction statistics. These descriptors reveal how bubble behavior varies with diameter $d$ and gas volume fraction $\varepsilon$, and form the basis for the evaluation metrics introduced in Section~\ref{sec:metrics}.

\textbf{Kinematics} We look at bubble centroid velocity magnitude $||\mathbf{v}(t)||=||\dot{ \mathbf{x}}(t)||$ to analyze the system's kinetic energy, we use the rise angle $\theta_z=\arccos(v_z(t) / ||\mathbf{v}(t)||)$, to quantify buoyancy-to-drag dominance and the tortuosity $\tau=\sum_{t=0}^{T}{||\dot{\mathbf{x}}(t)||} \ / \ ||\mathbf{x}(T)-\mathbf{x}(0)||$, which is a geometric property describing how "twisted" a bubble trajectory is, expressed by the ratio of arc length to displacement, to investigate movement efficiency.

\textbf{Morphology} Shape information of the bubble interfaces is expressed using three quantities, of which surface area $A$ is the first. The sphericity, $\Psi=\pi^{\frac{1}{3}}(6V)^{\frac{2}{3}} / A$, is the ratio of surface area $A$ of a sphere with the same volume $V$  to the object's surface area, measuring how closely the bubble's shape resembles a perfect sphere. We also look at the amount of change of the surface over time by transforming SH coefficients $\mathbf{c}$ back into point clouds and calculating the radial point displacements over time, while taking the mean over the displacements, to get surface velocity at each timestep, $\langle | \dot{r}(\theta, \phi) | \rangle$.  

\textbf{Interactions} The motion and deformation of bubbles in close proximity are strongly influenced by hydrodynamic interactions, primarily arising from drag forces \cite{Roghair2011ces}. We quantify spatial organization using the radial distribution function \(g(r)\), which describes how bubble density varies with distance from a reference bubble centroid. To characterize velocity interactions, we compute the angle between velocity vectors for bubble pairs within a distance of \(4d\), $\theta_{i,j}=\arccos(\mathbf{v}_i \cdot \mathbf{v}_j / ||\mathbf{v}_i|| ||\mathbf{v}_j||)$.

\begin{wrapfigure}[21]{R}{0.44\textwidth}
\adjustbox{trim=0.2cm 0.0cm 0cm 0.4cm}{
\includegraphics[width=1.01\linewidth]{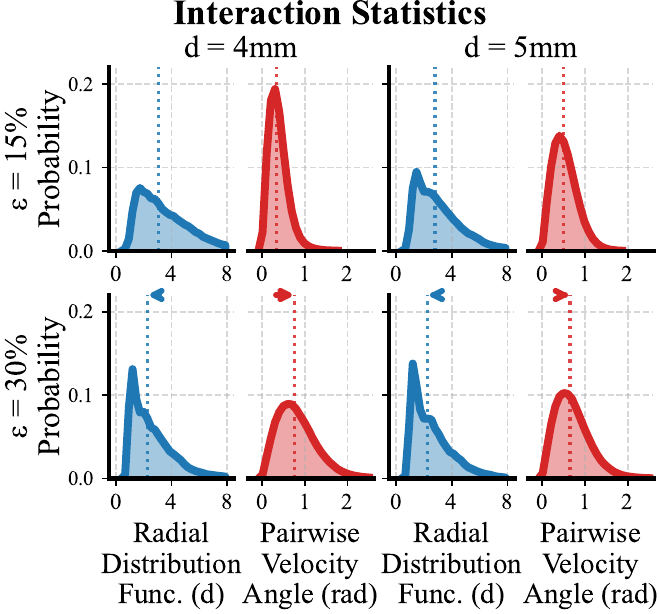}}
\vspace{-0.6em}
\caption{Probability mass functions of interaction quantities from Section \ref{statistics} for two diameters $d$ and gas fractions $\varepsilon$. Dotted line indicates the median, arrows indicate median shift, number of bins is 30, see Fig \ref{pmf1}.}
\label{pmf2}
\end{wrapfigure}

The empirical probability mass functions of the eight mentioned properties from dataset configurations $d\in \{4,5\}$ and $\varepsilon\in \{ 15,30 \}$ are plotted in Figures~\ref{pmf1} and \ref{pmf2}, using $d$ as unit of length. Looking at the densities and their medians in Figure~\ref{pmf1}, represented by the dotted lines, we see notable shifts from $\varepsilon=15\%$ to $\varepsilon=30\%$ across all descriptors. Bubbles are slower, rise at a less vertical angle and move in more tortuous paths when increasing $\varepsilon$ as a result of increased drag forces \cite{lockett1975bubbly_flow,bridge1963mechanics_I,simonnet2007liquid_mixing}. This difference is more apparent for smaller bubbles \cite{Roghair2011ces}. Examining the morphology densities on the right side of Figure \ref{pmf1}, we note that bubbles have lower surface area, are more spherical and have a higher rate of interface deformation at higher $\varepsilon$. These effects are again dampened at $d=5$. 
In terms of surface velocity, we notice an increase in deformation for $d=4$ which is not present at $d=5$ when raising gas density. The interaction statistics are shown in Figure \ref{pmf2}. A peak in bubble density at distance $r\approx2d$ can be observed across all regimes, while the median depends inversely on $\varepsilon$ as expected. Lastly, the pairwise angles between velocity vectors increases with $\varepsilon$, indicating that more streamlined upwards motion is replaced by inefficient horizontal motion.


\subsection{Evaluation Metrics}

\label{sec:metrics}
Measuring the performance of a model is often done using displacement errors in the field of trajectory forecasting and point-based distance metrics and volumetric measures in the scope of shape prediction. These standard metrics like Average Displacement Error, Chamfer Distance and Intersection over Union, penalize deviations from individual ground truth trajectories or shapes without regard for predicted collective physical behavior of the system. Distribution-based evaluation additionally measures whether a model captures the statistical structure of dynamically meaningful quantities, ensuring that predictions are physically consistent at the population level rather than merely geometrically close to specific realizations. We advocate that density-based metrics of relevant domain-informed quantities are crucial in a machine learning evaluation pipeline, especially for probabilistic generative models, designed to sample multiple different possible predictions. For stochastic, turbulent systems such as bubble swarms, where multiple physically valid futures exist for given initial conditions, standard metrics would conflate diversity with model error and therefore provide an incomplete picture of model quality.

As such, in addition to displacement and volumetric errors, we measure the similarity of the predicted and ground truth distributions $p,q$ using the 1-Wasserstein distance (Earth Mover's Distance): 
\begin{equation}
    \mathcal{W}_1(p,q)= \inf_{\gamma\in\Pi(p,q)} \mathbb{E}_{(x,y) \sim \gamma} \ |x-y|
\end{equation} 
where $\Pi(p,q)$ denotes the set of all couplings of $p$ and $q$. Unlike KL divergence, $\mathcal{W}_1$ is a proper metric and has the natural interpretation as the minimum cost of transporting one distribution to another, making it well suited to physical quantities that may be multimodal or heavy-tailed.

We apply this to the marginal distributions of the kinematic, morphological, and interaction quantities described in Section \ref{statistics}. For each quantity, we compute $\mathcal{W}_1$ between the empirical distribution of predicted and ground truth values aggregated across all bubbles and time steps. As these quantities have different scales, raw $\mathcal{W}_1$ distances are normalized by the interquartile range of the ground truth distributions before aggregation, yielding a dimensionless score per quantity, insensitive to heavy-tailed distributions. A scalar benchmark metric for model comparison can then be obtained by averaging across all normalized $\mathcal{W}_1$ distances. To anchor the reported scores, a reference baseline should always be included in the evaluation. 

In addition to density-based evaluation, we provide improvements on common point-based metrics measuring errors on positional and shape data. The Average/Final Displacement Error, ADE $=\langle  ||\mathbf{x}_i(t)-\hat{\mathbf{x}}_i(t)|| \rangle$, FDE $=\langle  ||\mathbf{x}_i(T)-\hat{\mathbf{x}}_i(T)|| \rangle$, depend on both prediction horizon and spatial scale, which makes comparison across different time scales and units of measurement harder. We propose normalizing the mean displacement errors by ground truth arc length, as Relative Average/Final Displacement Error, for a dimensionless and more interpretable position-based error metric. Defined as R-ADE $=\langle ||\mathbf{x}_i(t)-\hat{\mathbf{x}}_i(t)|| \ / \ \sum_{t=0}^T ||  \dot{\mathbf{x}}_i(t)|| \rangle$, it measures the mean ratio of displacement error to distance traveled. We extend this to evaluating deformation errors over time, as the Relative Average Chamfer Distance (R-ACD), which is normalized by the total change in $r(\theta, \phi)$ over time, as R-ACD $=\langle \ CD(r_i(t),\hat{r}_i(t)) \ / \ \sum_{t=0}^T \langle |\dot{r}_i(t)|\rangle \ \rangle$ with $CD(A,B)=\frac{1}{|A|}\sum_{a\in A}min_{b\in B}||a-b||+\frac{1}{|B|}\sum_{b\in B}min_{a\in A}||a-b||$ for bubble point clouds $A$ and $B$ \cite{DBLP:journals/corr/abs-2111-12702}. 







\section{Evaluation}
BubbleSH targets a gap in multiphase flow modeling, DNS resolve individual bubble dynamics with high fidelity but are computationally expensive, while practical large-scale Euler-Lagrange models collapse this rich per-bubble information into averaged correlations. Data-driven emulators trained on BubbleSH allow for bridging this gap, learning to reproduce coherent bubble dynamics at a fraction of the computational cost. To demonstrate this, along with our proposed metrics, we have trained a probabilistic generative forecasting model on BubbleSH that jointly predicts bubble trajectories and shape evolution.
\begin{figure}
\adjustbox{trim=0.7cm 0.3cm 0cm 0.8cm}{
    \includegraphics[width=1.04\textwidth]{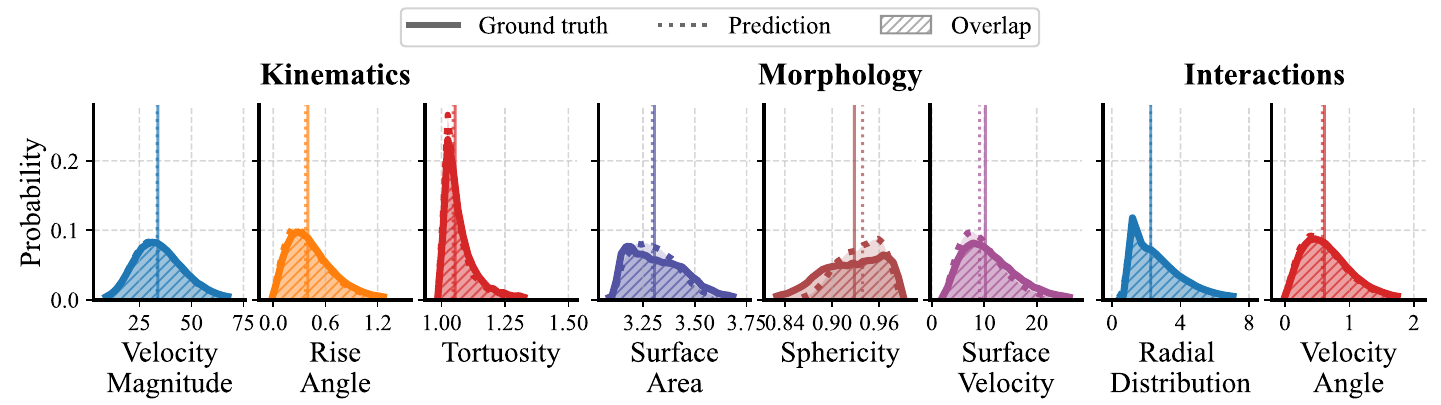}}
    \caption{Probability mass functions of predicted and ground truth trajectories from the Model $(L=5)$ experiment in Section \ref{exp} using quantities defined in Section \ref{statistics}.}
    \label{comp}
\end{figure}

\subsection{Model}
Concretely, we implemented a data-driven surrogate based on STFlow \cite{brinke2026}, a generative model using the conditional flow matching framework \cite{Liu2022FlowSA, lipman2023}. We frame trajectory simulation as learning the joint probability distribution over future states conditioned on observed initial frames. This formulation reflects the sensitivity of bubble-swarm dynamics to turbulent interactions and local perturbations, where multiple physically valid futures may exist for a given initial conditions. Our model constructs an informed prior $p_0$, where the observed part is kept fixed, while the unobserved future is initialized using a stochastic random-walk of all 32 bubbles over velocities $\mathbf{v} \in \mathbb{R}^{(T-T_c) \times 32 \times 3}$ and changes in spherical harmonic coefficients $\dot{\mathbf{c}} \in \mathbb{R}^{(T-T_c) \times 32 \times (L+1)^2}$ for observed trajectory length $T_c$. Positions and velocities are represented in the cylindrical coordinate system to align with the translational and vertical-axis rotational symmetry of the dynamics. Only velocity, acceleration and relative position information is used to represent the centroid trajectories, making the emulator equivariant to translation.

The architecture consists of alternating spatial and temporal processing layers. Our permutation-equivariant spatial message-passing graph neural network layer is based on Neural Message Passing \cite{pmlr-v70-gilmer17a} and captures multi-bubble interactions by aggregating messages along edges defined by inter-bubble distances, operating independently at each time step. The temporal layer, implemented as a 1D UNet, models the dynamics of individual trajectories across time. Both layers share high-dimensional latent node-level embeddings as part of a spatiotemporal graph, which are updated after every layer. Separate decoders for $\mathbf{v}$ and each  SH order within $\dot{\mathbf{c}}$ produce the outputs that denoise the prior $p_0$ into targets. Additional model details can be found in Appendix \ref{appendix:algos}, \ref{appendix:mp_layer} and  \ref{appendix:hyperparam}.

\subsection{Experiments} \label{exp}
We evaluate the model on the presented dataset and study its performance using both the point displacement and distributional-based metrics discussed in Section \ref{sec:metrics}.

\textbf{Data} The model is trained and evaluated on a large subset of BubbleSH spanning multiple diameters and gas fractions. All trajectories are sampled at a temporal resolution of \(10^{-3}\, \mathrm{s}\), providing a consistent setting across configurations. Fixed length trajectory windows of 60 ms and $T=30$ frames are extracted using a sliding-window approach. For each window, the first 20 ms ($T_c=10$) are used as the observed conditioning prefix, and the remaining portion are treated as the prediction target. The dataset is split into training, validation, and test sets with proportions of $80\%$, $5\%$, and $15\%$, respectively, containing $2094$ samples.

\textbf{Metrics} Following Section~\ref{sec:metrics}, for centroid position accuracy, we report the Relative ADE and FDE metrics. For assessment of the bubble deformations, we report the Intersection over Union and Relative Average Chamfer Distance. To assess whether the model reproduces physically meaningful dynamics, we also report the normalized 1-Wasserstein distance between predicted and ground-truth marginal distributions of the kinematic, morphological, and interaction quantities.

\begin{table}[t!]
\vspace{-1em}
\caption{Experimental study comparing the prior baseline, our generative model with $L=5$, $L=3$, and a variant with conditional independence between bubbles. Metrics follow Section~\ref{statistics}.}
\centering
\small
\setlength{\tabcolsep}{3pt}
\adjustbox{trim=0.0cm 0.5cm 0cm 0.0cm}{
\begin{tabular}{lcccccccc}
\toprule
 & R-ADE $\downarrow$ & R-FDE $\downarrow$ & IoU $\uparrow$ & R-ACD $\downarrow$ & Kinematics $\downarrow$ & Morphology $\downarrow$ & Interaction $\downarrow$&  \\ 
\midrule
Prior $p_0$ & 0.283 & 0.540 & 0.583 & 0.945 & 0.589 & 0.642 & 0.0036 \\ 
Model \footnotesize{$(L=5)$} & 0.185 & 0.322 & \textbf{0.778} & \textbf{0.442} & \textbf{0.131} & 0.209 & \textbf{0.0010} \\ 
 \ \ \textit{w/ independence} & 0.188 & 0.328 & 0.774 & 0.460 & 0.132 & \textbf{0.191} & 0.0013 \\ 
Model \footnotesize{$(L=3)$} & \textbf{0.163} &\textbf{0.295} & 0.777 & 0.444 & 0.134 & 0.340 & 0.0011 \\ 
\bottomrule \\
\end{tabular} }

\label{results} 
\end{table}

We compare four configurations: (i) a baseline where the generated trajectory is sampled from the random walk prior $p_0$ itself; (ii) the presented model using $L=5,(5+1)^2=36$ SH coefficients; (iii) a variant where the bubbles are modeled as being conditionally independent by removing the spatial interaction layers and (iiii) the model using $L=3,(3+1)^2=16$ SH coefficients.

\begin{wraptable}[10]{R}{0.35\textwidth}
\centering
\footnotesize
\setlength{\tabcolsep}{4pt}
\caption{Comparison of required time to run 0.1s of simulation.}
\label{table:speed}
\vspace{0.7em}
\begin{tabular}{lccc}
    \toprule
     & $\varepsilon$ & Time& Speedup \\
    \midrule
    FT & $5$ & $2.6$ days & $1\times$ \\
    FT & $40$  & $0.6$ days & \footnotesize $1\times$ \\
    Emulator & \footnotesize$5$ & $0.85$ sec  & \footnotesize$260$,$000\times$ \\
    Emulator & \footnotesize$40$ & $0.85$ sec  & \footnotesize$58$,$000\times$ \\
    \bottomrule
\end{tabular}
\end{wraptable}The results in Table \ref{results} demonstrate that the full generative model with $L=5$ achieves the strongest overall performance, improving over the random walk prior across all metrics. In Figure \ref{comp} we plot the PMF's calculated from predictions and targets of the $L=5$ experiment and highlight their overlap, from which we qualitatively confirm that the learned dynamics are aligned with the data. The distributional Wasserstein scores show substantial improvement over the prior, indicating that the model captures realistic behavior at the population level rather than memorizing individual trajectories. The ablation with conditional independence performs surprisingly similar to the full model, indicating that the model does not yet learn to capture spatial inter-bubble interactions well. The $L=3$ model produces the lowest displacement errors but higher morphology dissimilarity, confirming that truncating the spherical harmonics too aggressively limits the model's ability to reproduce accurate deformations. Beyond predictive accuracy, we compare the speedup in simulation time by using the generative model in Table \ref{table:speed}. The trained emulator achieves four to five orders-of-magnitude speedup over the FT simulator. This speedup, combined with the physically grounded evaluation framework, positions BubbleSH as practical stepping stone towards scalable data-driven multiphase flow modeling that retains per-bubble fidelity.

\section{Conclusion}
\label{sec:conclusion}

We presented BubbleSH, a dataset of transient, three-dimensional bubble swarm dynamics derived from high-fidelity interface-resolving direct numerical simulation. The dataset pairs time-resolved centroid trajectories with continuous spherical harmonic shape descriptors across multiple bubble sizes and gas volume fractions. BubbleSH sits between rigid particle dynamics benchmarks and mesh-based simulation benchmarks; the bubbles interact collectively like point particles but carry evolving shape states.
As demonstrated by our generative baseline, models trained on BubbleSH can reproduce coherent bubble dynamics at four to five orders of magnitude lower computational cost than the underlying DNS, offering a practical path toward predictive swarm-scale modeling that preserves individual bubble variability. The accompanying evaluation framework provides both point-wise and distribution-based metrics for assessing trajectory accuracy and physical consistency. Together, these provide a foundation for developing and comparing data-driven multiphase flow emulators that must jointly handle complex many-body interactions and high-resolution deformable geometry.


\textbf{Limitations and Future Work}
By simulating in a fully periodic domain, the representation of large-scale inhomogeneities, wall effects, and long-range plume dynamics is restricted. The simulations are idealized air-water systems without breakup, coalescence, surfactants, or contaminants, and therefore do not capture physicochemical effects that may be important in industrial applications. The current parameter space covers strongly deformable bubbles in water at selected \(\varepsilon\) values, corresponding roughly to Reynolds numbers of 100-1000 and E\"otv\"os numbers of 2-5, but excludes nearly spherical, weakly deformed ellipsoidal, and spherical-cap regimes. Although spherical harmonics provide a compact shape representation, truncation can exhibit high-frequency Gibbs-type oscillations that may affect local geometric quantities. While our initial generative benchmark highlights the use of BubbleSH, the model remains limited in its treatment of symmetry and long-horizon stability. Future work could incorporate the rotational structure of spherical harmonics and study more stable rollout methods beyond fixed-window forecasting.



\bibliographystyle{plainnat}
\bibliography{bibliography}

@article{aganinDynamicsGasBubbles2022,
  title = {Dynamics of {{Gas Bubbles}} in a {{Spherical Cluster Under}} a {{Single Harmonic Pulse}} of {{Liquid Compression}}},
  author = {Aganin, I. A.},
  year = 2022,
  month = may,
  journal = {Lobachevskii Journal of Mathematics},
  volume = {43},
  number = {5},
  pages = {1057--1063},
  issn = {1995-0802, 1818-9962},
  doi = {10.1134/S1995080222080030},
  urldate = {2026-04-10},
  langid = {english}
}

@incollection{brechbuhlerRepresentation3DShape1992a,
  title = {Towards {{Representation}} of {{3D Shape}}: {{Global Surface Parametrization}}},
  shorttitle = {Towards {{Representation}} of {{3D Shape}}},
  booktitle = {Visual {{Form}}},
  author = {Brechb{\"u}hler, {\relax Ch}. and Gerig, G. and K{\"u}bler, O.},
  editor = {Arcelli, Carlo and Cordella, Luigi P. and Di Baja, Gabriella Sanniti},
  year = 1992,
  pages = {79--88},
  publisher = {Springer US},
  address = {Boston, MA},
  doi = {10.1007/978-1-4899-0715-8_9},
  urldate = {2026-04-10},
  isbn = {978-1-4899-0717-2 978-1-4899-0715-8},
  langid = {english}
}

@article{brennShapeOscillationsPath2006,
  title = {Shape Oscillations and Path Transition of Bubbles Rising in a Model Bubble Column},
  author = {Brenn, G. and Kolobaric, V. and Durst, F.},
  year = 2006,
  month = jun,
  journal = {Chemical Engineering Science},
  volume = {61},
  number = {12},
  pages = {3795--3805},
  issn = {00092509},
  doi = {10.1016/j.ces.2005.12.016},
  urldate = {2026-01-05},
  langid = {english}
}

@article{cattaneoShapeModesJet2025,
  title = {Shape Modes and Jet Formation on Ultrasound-Driven Wall-Attached Bubbles},
  author = {Cattaneo, Marco and Presse, Louan and Shakya, Gazendra and Renggli, Thomas and Luki{\'c}, Bratislav and Prasanna, Anunay and Meyer, Daniel Werner and Rack, Alexander and Supponen, Outi},
  year = 2025,
  month = aug,
  journal = {Journal of Fluid Mechanics},
  volume = {1017},
  pages = {A9},
  issn = {0022-1120, 1469-7645},
  doi = {10.1017/jfm.2025.10457},
  urldate = {2026-04-10},
  langid = {english}
}

@article{inserraMathematicalModellingAcoustic2025,
  title = {Mathematical Modelling for Acoustic Microstreaming Produced by a Gas Bubble Undergoing Asymmetric Oscillations},
  author = {Inserra, Claude and Mauger, Cyril and {Blanc-Benon}, Philippe and Doinikov, Alexander A.},
  year = 2025,
  month = may,
  journal = {Journal of Fluid Mechanics},
  volume = {1010},
  pages = {A48},
  issn = {0022-1120, 1469-7645},
  doi = {10.1017/jfm.2025.284},
  urldate = {2026-04-10},
  langid = {english}
}

@article{nortjeSphericalHarmonicsSurface2015a,
  title = {Spherical {{Harmonics}} for {{Surface Parametrisation}} and {{Remeshing}}},
  author = {Nortje, Caitlin R. and Ward, Wil O. C. and Neuman, Bartosz P. and Bai, Li},
  year = 2015,
  journal = {Mathematical Problems in Engineering},
  volume = {2015},
  number = {1},
  pages = {582870},
  issn = {1563-5147},
  doi = {10.1155/2015/582870},
  urldate = {2026-04-10},
  langid = {english}
}

@inproceedings{shenLargeScaleModelingParametric2006a,
  title = {Large-{{Scale Modeling}} of {{Parametric Surfaces Using Spherical Harmonics}}},
  booktitle = {Third {{International Symposium}} on {{3D Data Processing}}, {{Visualization}}, and {{Transmission}} ({{3DPVT}}'06)},
  author = {Shen, Li and Chung, Moo K.},
  year = 2006,
  month = jun,
  pages = {294--301},
  publisher = {IEEE},
  address = {Chapel Hill, NC, USA},
  doi = {10.1109/3DPVT.2006.86},
  urldate = {2026-04-10},
  isbn = {978-0-7695-2825-0},
  langid = {english}
}

@article{yu3DBubblesExperimentalDataset2026,
  title = {{{3DBubbles}}: {{An}} Experimental Dataset for Model Training and Benchmarking},
  shorttitle = {{{3DBubbles}}},
  author = {Yu, Baodi and Chen, Qian and Qin, Yanwei and Wang, Sunyang and Su, Xiaohui and Meng, Fanyong},
  year = 2026,
  journal = {AIChE Journal},
  volume = {72},
  number = {1},
  pages = {e70029},
  issn = {1547-5905},
  doi = {10.1002/aic.70029},
  urldate = {2026-04-10},
  langid = {english}
}

@article{zhou3DSurfaceFiltering2004,
  title = {{{3D}} Surface Filtering Using Spherical Harmonics},
  author = {Zhou, Kun and Bao, Hujun and Shi, Jiaoying},
  year = 2004,
  month = feb,
  journal = {Computer-Aided Design},
  volume = {36},
  number = {4},
  pages = {363--375},
  issn = {00104485},
  doi = {10.1016/S0010-4485(03)00098-8},
  urldate = {2026-04-10},
  copyright = {https://www.elsevier.com/tdm/userlicense/1.0/},
  langid = {english}
}

@article{lalanneNonlinearShapeOscillations2015,
  title = {Non-Linear Shape Oscillations of Rising Drops and Bubbles: {{Experiments}} and Simulations},
  shorttitle = {Non-Linear Shape Oscillations of Rising Drops and Bubbles},
  author = {Lalanne, Benjamin and Abi Chebel, Nicolas and Vejra{\v z}ka, Ji{\v r}{\'i} and Tanguy, S{\'e}bastien and Masbernat, Olivier and Risso, Fr{\'e}d{\'e}ric},
  year = 2015,
  month = dec,
  journal = {Physics of Fluids},
  volume = {27},
  number = {12},
  pages = {123305},
  issn = {1070-6631, 1089-7666},
  doi = {10.1063/1.4936980},
  urldate = {2026-04-14},
  langid = {english}
}

@article{Bunner2002a,
   author = {Bernard Bunner and Grétar Tryggvason},
   doi = {10.1017/S0022112002001179},
   issn = {00221120},
   journal = {Journal of Fluid Mechanics},
   month = {9},
   pages = {17-52},
   publisher = {Cambridge University Press},
   title = {Dynamics of homogeneous bubbly flows part 1. Rise velocity and microstructure of the bubbles},
   volume = {466},
   year = {2002}
}

@article{Bunner2002b,
   author = {Bernard Bunner and Grétar Tryggvason},
   doi = {10.1017/S0022112002001180},
   journal = {Journal of Fluid Mechanics},
   pages = {53-84},
   title = {Dynamics of homogeneous bubbly flows Part 2. Velocity fluctuations},
   volume = {466},
   year = {2002}
}

@article{Unverdi1992,
   author = {Salih Ozen Unverdi and Grétar Tryggvason},
   doi = {https://doi.org/10.1016/0021-9991(92)90307-K},
   issn = {0021-9991},
   issue = {1},
   journal = {Journal of Computational Physics},
   pages = {25-37},
   title = {A front-tracking method for viscous, incompressible, multi-fluid flows},
   volume = {100},
   url = {https://www.sciencedirect.com/science/article/pii/002199919290307K},
   year = {1992}
}

@article{Roghair2016,
   author = {I. Roghair and M. Van Sint Annaland and J. A.M. Kuipers},
   doi = {10.1016/j.ces.2016.06.026},
   issn = {00092509},
   journal = {Chemical Engineering Science},
   month = {10},
   pages = {351-369},
   publisher = {Elsevier Ltd},
   title = {An improved Front-Tracking technique for the simulation of mass transfer in dense bubbly flows},
   volume = {152},
   year = {2016}
}

@article{Roghair2011ijmf,
   author = {Ivo Roghair and Julián Martínez Mercado and Martin Van Sint Annaland and Hans Kuipers and Chao Sun and Detlef Lohse},
   doi = {10.1016/j.ijmultiphaseflow.2011.07.004},
   issn = {03019322},
   issue = {9},
   journal = {International Journal of Multiphase Flow},
   month = {11},
   pages = {1093-1098},
   title = {Energy spectra and bubble velocity distributions in pseudo-turbulence: Numerical simulations vs. experiments},
   volume = {37},
   year = {2011}
}

@article{Tagawa2013,
   author = {Yoshiyuki Tagawa and Ivo Roghair and Vivek N. Prakash and Martin Van Sint Annaland and Hans Kuipers and Chao Sun and Detlef Lohse},
   doi = {10.1017/jfm.2013.100},
   issn = {14697645},
   journal = {Journal of Fluid Mechanics},
   publisher = {Cambridge University Press},
   title = {The clustering morphology of freely rising deformable bubbles},
   volume = {721},
   year = {2013}
}

@article{Roghair2013aiche,
   author = {Ivo Roghair and Martin Van Sint Annaland and Hans J.A.M. Kuipers},
   doi = {10.1002/aic.13949},
   issn = {00011541},
   issue = {5},
   journal = {AIChE Journal},
   month = {5},
   pages = {1791-1800},
   title = {Drag force and clustering in bubble swarms},
   volume = {59},
   year = {2013}
}

@article{Dijkhuizen2010a,
   author = {W. Dijkhuizen and I. Roghair and M. Van Sint Annaland and J. A.M. Kuipers},
   doi = {10.1016/j.ces.2009.10.022},
   issn = {00092509},
   issue = {4},
   journal = {Chemical Engineering Science},
   pages = {1427-1437},
   title = {DNS of gas bubbles behaviour using an improved 3D front tracking model-Model development},
   volume = {65},
   year = {2010}
}

@article{Dijkhuizen2010b,
   author = {W. Dijkhuizen and I. Roghair and M. Van Sint Annaland and J. A.M. Kuipers},
   doi = {10.1016/j.ces.2009.10.021},
   issn = {00092509},
   issue = {4},
   journal = {Chemical Engineering Science},
   pages = {1415-1426},
   title = {DNS of gas bubbles behaviour using an improved 3D front tracking model-Drag force on isolated bubbles and comparison with experiments},
   volume = {65},
   year = {2010}
}

@article{Roghair2011ces,
   author = {I. Roghair and Y. M. Lau and N. G. Deen and H. M. Slagter and M. W. Baltussen and M. Van Sint Annaland and J. A.M. Kuipers},
   doi = {10.1016/j.ces.2011.02.030},
   issn = {00092509},
   issue = {14},
   journal = {Chemical Engineering Science},
   month = {7},
   pages = {3204-3211},
   title = {On the drag force of bubbles in bubble swarms at intermediate and high Reynolds numbers},
   volume = {66},
   year = {2011}
}

@incollection{Youngs1982,
  author    = {Youngs, D. L.},
  title     = {Time-Dependent Multi-material Flow with Large Fluid Distortion},
  booktitle = {Numerical Methods in Fluid Dynamics},
  editor    = {Morton, K. W. and Baines, M. J.},
  publisher = {Academic Press},
  year      = {1982},
  address   = {London},
  month     = jan,
  pages     = {273--285}
}

@article{Rider1998,
   author = {William J Rider and Douglas B Kothe},
   doi = {https://doi.org/10.1006/jcph.1998.5906},
   issn = {0021-9991},
   issue = {2},
   journal = {Journal of Computational Physics},
   pages = {112-152},
   title = {Reconstructing Volume Tracking},
   volume = {141},
   url = {https://www.sciencedirect.com/science/article/pii/S002199919895906X},
   year = {1998}
}

@article{Hirt1981,
   author = {C W Hirt and B D Nichols},
   doi = {https://doi.org/10.1016/0021-9991(81)90145-5},
   issn = {0021-9991},
   issue = {1},
   journal = {Journal of Computational Physics},
   pages = {201-225},
   title = {Volume of fluid (VOF) method for the dynamics of free boundaries},
   volume = {39},
   url = {https://www.sciencedirect.com/science/article/pii/0021999181901455},
   year = {1981}
}

@article{SintAnnaland2006,
author = {van Sint Annaland, M. and Dijkhuizen, W. and Deen, N. G. and Kuipers, J. A. M.},
title = {Numerical simulation of behavior of gas bubbles using a 3-D front-tracking method},
journal = {AIChE Journal},
volume = {52},
number = {1},
pages = {99-110},
doi = {https://doi.org/10.1002/aic.10607},
url = {https://aiche.onlinelibrary.wiley.com/doi/abs/10.1002/aic.10607},
eprint = {https://aiche.onlinelibrary.wiley.com/doi/pdf/10.1002/aic.10607},
year = {2006}
}

@article{SintAnnaland2005,
title = {Numerical simulation of gas bubbles behaviour using a three-dimensional volume of fluid method},
journal = {Chemical Engineering Science},
volume = {60},
number = {11},
pages = {2999-3011},
year = {2005},
issn = {0009-2509},
doi = {https://doi.org/10.1016/j.ces.2005.01.031},
url = {https://www.sciencedirect.com/science/article/pii/S0009250905000564},
author = {M. {van Sint Annaland} and N.G. Deen and J.A.M. Kuipers}
}

@article{Darmana2006,
author = {Darmana, D. and Deen, N. G. and Kuipers, J. A. M.},
title = {Detailed 3D Modeling of Mass Transfer Processes in Two-Phase Flows with Dynamic Interfaces},
journal = {Chemical Engineering \& Technology},
volume = {29},
number = {9},
pages = {1027-1033},
doi = {https://doi.org/10.1002/ceat.200600156},
url = {https://onlinelibrary.wiley.com/doi/abs/10.1002/ceat.200600156},
eprint = {https://onlinelibrary.wiley.com/doi/pdf/10.1002/ceat.200600156},
year = {2006}
}

@article {kimDeepLearningBasedAutomated2021,
	Title = {Deep learning-based automated and universal bubble detection and mask extraction in complex two-phase flows},
	Author = {Kim, Yewon and Park, Hyungmin},
	DOI = {10.1038/s41598-021-88334-0},
	Number = {1},
	Volume = {11},
	Month = {April},
	Year = {2021},
	Journal = {Scientific reports},
	ISSN = {2045-2322},
	Pages = {8940},
	URL = {https://europepmc.org/articles/PMC8076184},
}

@article{BubGAN,
title = {BubGAN: Bubble generative adversarial networks for synthesizing realistic bubbly flow images},
journal = {Chemical Engineering Science},
volume = {204},
pages = {35-47},
year = {2019},
issn = {0009-2509},
doi = {https://doi.org/10.1016/j.ces.2019.04.004},
url = {https://www.sciencedirect.com/science/article/pii/S0009250919303677},
author = {Yucheng Fu and Yang Liu}
}

@article{Soibam,
title = {Application of deep learning for segmentation of bubble dynamics in subcooled boiling},
journal = {International Journal of Multiphase Flow},
volume = {169},
pages = {104589},
year = {2023},
issn = {0301-9322},
doi = {https://doi.org/10.1016/j.ijmultiphaseflow.2023.104589},
url = {https://www.sciencedirect.com/science/article/pii/S0301932223002094},
author = {Jerol Soibam and Valentin Scheiff and Ioanna Aslanidou and Konstantinos Kyprianidis and Rebei {Bel Fdhila}}
}

@article{Hessenkemper3D,
title = {3D detection and tracking of deformable bubbles in swarms with the aid of deep learning models},
journal = {International Journal of Multiphase Flow},
volume = {179},
pages = {104932},
year = {2024},
issn = {0301-9322},
doi = {https://doi.org/10.1016/j.ijmultiphaseflow.2024.104932},
url = {https://www.sciencedirect.com/science/article/pii/S030193222400209X},
author = {Hendrik Hessenkemper and Lantian Wang and Dirk Lucas and Shiyong Tan and Rui Ni and Tian Ma}
}

@InProceedings{kipf2018,
  title = 	 {Neural Relational Inference for Interacting Systems},
  author =       {Kipf, Thomas and Fetaya, Ethan and Wang, Kuan-Chieh and Welling, Max and Zemel, Richard},
  booktitle = 	 {Proceedings of the 35th International Conference on Machine Learning},
  pages = 	 {2688--2697},
  year = 	 {2018},
  editor = 	 {Dy, Jennifer and Krause, Andreas},
  volume = 	 {80},
  series = 	 {Proceedings of Machine Learning Research},
  month = 	 {10--15 Jul},
  publisher =    {PMLR},
  url = 	 {https://proceedings.mlr.press/v80/kipf18a.html}
}

@InProceedings{egnnSatorras,
  title = 	 {E(n) Equivariant Graph Neural Networks},
  author =       {Satorras, V\'{\i}ctor Garcia and Hoogeboom, Emiel and Welling, Max},
  booktitle = 	 {Proceedings of the 38th International Conference on Machine Learning},
  pages = 	 {9323--9332},
  year = 	 {2021},
  editor = 	 {Meila, Marina and Zhang, Tong},
  volume = 	 {139},
  series = 	 {Proceedings of Machine Learning Research},
  month = 	 {18--24 Jul},
  publisher =    {PMLR},
  url = 	 {https://proceedings.mlr.press/v139/satorras21a.html}
}

@InProceedings{GNSSanchez_Gonzalez,
  title = 	 {Learning to Simulate Complex Physics with Graph Networks},
  author =       {Sanchez-Gonzalez, Alvaro and Godwin, Jonathan and Pfaff, Tobias and Ying, Rex and Leskovec, Jure and Battaglia, Peter},
  booktitle = 	 {Proceedings of the 37th International Conference on Machine Learning},
  pages = 	 {8459--8468},
  year = 	 {2020},
  editor = 	 {III, Hal Daumé and Singh, Aarti},
  volume = 	 {119},
  series = 	 {Proceedings of Machine Learning Research},
  month = 	 {13--18 Jul},
  publisher =    {PMLR},
  url = 	 {https://proceedings.mlr.press/v119/sanchez-gonzalez20a.html}
}

@article{MeshGraphNets,
  author       = {Tobias Pfaff and
                  Meire Fortunato and
                  Alvaro Sanchez{-}Gonzalez and
                  Peter W. Battaglia},
  title        = {Learning Mesh-Based Simulation with Graph Networks},
  journal      = {CoRR},
  volume       = {abs/2010.03409},
  year         = {2020},
  url          = {https://arxiv.org/abs/2010.03409},
  eprinttype   = {arXiv},
  eprint       = {2010.03409},
  bibsource    = {dblp computer science bibliography, https://dblp.org}
}

@inproceedings{LagrangeBench,
 author = {Toshev, Artur and Galletti, Gianluca and Fritz, Fabian and Adami, Stefan and Adams, Nikolaus},
 booktitle = {Advances in Neural Information Processing Systems},
 editor = {A. Oh and T. Naumann and A. Globerson and K. Saenko and M. Hardt and S. Levine},
 pages = {64857--64884},
 publisher = {Curran Associates, Inc.},
 title = {LagrangeBench: A Lagrangian Fluid Mechanics Benchmarking Suite},
 url = {https://proceedings.neurips.cc/paper_files/paper/2023/file/ccac3b120c7dc86d45f56830732b62be-Paper-Datasets_and_Benchmarks.pdf},
 volume = {36},
 year = {2023}
}

@misc{hendriks2026,
      title={Equivariant Flow Matching for Symmetry-Breaking Bifurcation Problems}, 
      author={Fleur Hendriks and Ondřej Rokoš and Martin Doškář and Marc G. D. Geers and Vlado Menkovski},
      year={2026},
      eprint={2509.03340},
      archivePrefix={arXiv},
      primaryClass={cs.LG},
      url={https://arxiv.org/abs/2509.03340}, 
}

@misc{brinke2026,
      title={STFlow: Data-Coupled Flow Matching for Geometric Trajectory Simulation}, 
      author={Kiet Bennema ten Brinke and Koen Minartz and Vlado Menkovski},
      year={2026},
      eprint={2505.18647},
      archivePrefix={arXiv},
      primaryClass={cs.LG},
      url={https://arxiv.org/abs/2505.18647}, 
}

@misc{lipman2023,
      title={Flow Matching for Generative Modeling}, 
      author={Yaron Lipman and Ricky T. Q. Chen and Heli Ben-Hamu and Maximilian Nickel and Matt Le},
      year={2023},
      eprint={2210.02747},
      archivePrefix={arXiv},
      primaryClass={cs.LG},
      url={https://arxiv.org/abs/2210.02747}, 
}

@article{bridge1963mechanics_I,
  author  = {Bridge, A. G. and Lapidus, L. and Elgin, J. C.},
  title   = {The Mechanics of Vertical Gas-Liquid Fluidized Systems. I. Countercurrent Flow},
  journal = {AIChE Journal},
  year    = {1963},
  volume  = {10},
  number  = {6},
  pages   = {819-826},
  doi     = {10.1002/aic.690090603}
}

@article{simonnet2007liquid_mixing,
  author  = {Simonnet, M. and Gentric, C. and Olmos, E.},
  title   = {Experimental determination of the drag coefficient in a swarm of bubbles},
  journal = {Chemical Engineering Science},
  year    = {2007},
  volume  = {62},
  number  = {3},
  pages   = {858--866},
  doi     = {10.1016/j.ces.2006.10.012}
}

@article{lockett1975bubbly_flow,
  author  = {Lockett, M. J. and Kirkpatrick, R. D.},
  title   = {Ideal Bubbly Flow and Actual Flow in Bubble Columns},
  journal = {Transactions of the Institution of Chemical Engineers},
  year    = {1975},
  volume  = {53},
  pages   = {267--273}
}

@article{Liu2022FlowSA,
  title={Flow Straight and Fast: Learning to Generate and Transfer Data with Rectified Flow},
  author={Xingchao Liu and Chengyue Gong and Qiang Liu},
  journal={ArXiv},
  year={2022},
  volume={abs/2209.03003},
  url={https://api.semanticscholar.org/CorpusID:252111177}
}

@InProceedings{pmlr-v70-gilmer17a,
  title = 	 {Neural Message Passing for Quantum Chemistry},
  author =       {Justin Gilmer and Samuel S. Schoenholz and Patrick F. Riley and Oriol Vinyals and George E. Dahl},
  booktitle = 	 {Proceedings of the 34th International Conference on Machine Learning},
  pages = 	 {1263--1272},
  year = 	 {2017},
  editor = 	 {Precup, Doina and Teh, Yee Whye},
  volume = 	 {70},
  series = 	 {Proceedings of Machine Learning Research},
  month = 	 {06--11 Aug},
  publisher =    {PMLR},
  url = 	 {https://proceedings.mlr.press/v70/gilmer17a.html}
}

@article{DBLP:journals/corr/abs-2111-12702,
  author       = {Tong Wu and
                  Liang Pan and
                  Junzhe Zhang and
                  Tai Wang and
                  Ziwei Liu and
                  Dahua Lin},
  title        = {Density-aware Chamfer Distance as a Comprehensive Metric for Point
                  Cloud Completion},
  journal      = {CoRR},
  volume       = {abs/2111.12702},
  year         = {2021},
  url          = {https://arxiv.org/abs/2111.12702},
  eprinttype   = {arXiv},
  eprint       = {2111.12702},
  bibsource    = {dblp computer science bibliography, https://dblp.org}
}

@inproceedings{dhariwal2021diffusionmodelsbeatgans,
  author        = {Prafulla Dhariwal and
                   Alexander Quinn Nichol},
  bibsource     = {dblp computer science bibliography, https://dblp.org},
  booktitle     = {Advances in Neural Information Processing Systems 34: Annual Conference
                   on Neural Information Processing Systems 2021, NeurIPS 2021, December
                   6-14, 2021, virtual},
  editor        = {Marc'Aurelio Ranzato and
                   Alina Beygelzimer and
                   Yann N. Dauphin and
                   Percy Liang and
                   Jennifer Wortman Vaughan},
  pages         = {8780--8794},
  title         = {{D}iffusion {M}odels {B}eat {G}{A}{N}s on {I}mage {S}ynthesis},
  url           = {https://proceedings.neurips.cc/paper/2021/hash/49ad23d1ec9fa4bd8d77d02681df5cfa-Abstract.html},
  year          = {2021}
}

@misc{sph_bubble_convert,
  author       = {Ivo Roghair and Douwe Orij and Rachna Ramesh},
  title        = {sph-bubble-convert},
  year         = {2026},
  howpublished = {\url{https://github.com/iroghair/sph-bubble-convert}},
  note         = {GitHub repository, commit e182c11, 2026-04-09}
}

\newpage


\appendix

\section{Technical appendices and supplementary material}

\subsection{Model Training and Inference Procedures} \label{appendix:algos}
The training and inference algorithms, inspired from \cite{brinke2026} are denoted below in Algorithm \ref{algo1} and Algorithm \ref{algo2}.

\begin{algorithm}[]

\caption{Conditional Flow Matching Training}
\label{algo1}
\centering
\begin{algorithmic}

\REQUIRE Train Dataset $p_1$, Model $v_\theta(\mathbf{x}, \mathbf{h}, \mathcal{E}, \tau)$
\WHILE{Training}
\STATE $\mathbf{x}_1 \sim p_1, \ \tau \sim \mathcal{U}[0,1]$ \hfill \# Sample batch $\mathbf{x}_1$ and denoising level $\tau$
\STATE $\mathbf{x}_0^{0:c} \gets \mathbf{x}_1^{0:c}$, \ $\mathbf{x}_0^{c:T} \gets \zeta(\mathbf{x}_1^{0:c})$
 \hfill \#  Make informed prior $\mathbf{x}_0$ from observed frames $\mathbf{x_1^{0:c}}$
\STATE $\mathbf{x}_\tau \sim p_\tau = \mathcal{N}(\tau \mathbf{x}_1 + (1-\tau) \mathbf{x}_0, \sigma_p^2)$  \hfill \# Sample trajectories from Gaussian probability path
\STATE $u \gets \mathbf{x}_1 - \mathbf{x}_0$  \hfill \# Calculate true vector field
\STATE $\mathbf{h} \gets \text{Features}(\mathbf{x}_\tau), \ \mathcal{E} \gets \text{Connectivity}(\mathbf{x}_\tau)$  \hfill \# Construct geometric trajectory
\STATE $v_\theta \gets v_\theta(\mathbf{x}_\tau, \mathbf{h}, \mathcal{E}, \tau)$   \hfill \# Forward pass
\STATE $\mathcal{L}_{CFM}(\theta) \gets \| v_\theta - u \|^2$  \hfill \# Calculate loss
\STATE $\theta \gets \theta + \alpha \nabla_\theta \mathcal{L}_{CFM}(\theta)$  \hfill \# Backpropagate and optimize
\ENDWHILE
\end{algorithmic}
\end{algorithm}

\begin{algorithm}

\caption{Inference using Euler Integration}
\label{algo2}
\begin{algorithmic}
\REQUIRE Test Dataset $p_1$, Trained model $v_\theta(\mathbf{x}, \mathbf{h}, \mathcal{E}, \tau)$, Number of Function Evaluations $n$
\FOR{$\mathbf{x}_1 \in p_1$}
\STATE $\mathbf{x}_0^{0:c} \gets \mathbf{x}_1^{0:c}$, \ $\mathbf{x}_0^{c:T} \gets \zeta(\mathbf{x}_1^{0:c})$
 \hfill \# Make informed prior $\mathbf{x}_0$ from observed frames $\mathbf{x_1^{0:c}}$
\STATE $\hat{\mathbf{x}} \gets \mathbf{x}_0$, \ $\Delta \tau \gets \frac{1}{n}$
\FOR{$\tau \in \{0, \frac{1}{n},...,\frac{n-1}{n}\}$}
\STATE $\mathbf{h} \gets \text{Features}(\hat{\mathbf{x}}), \ \mathcal{E} \gets \text{Connectivity}(\hat{\mathbf{x}})$  \hfill \# Construct geometric trajectory
\STATE $v_\theta \gets v_\theta(\hat{\mathbf{x}}, \mathbf{h}, \mathcal{E}, \tau)$   \hfill \# Forward pass
\STATE $\hat{\mathbf{x}} \gets \hat{\mathbf{x}} \ + \ \Delta \tau \cdot v_\theta$
 \hfill \# Update $\hat{\mathbf{x}}$ using Euler method
\ENDFOR
\ENDFOR
\end{algorithmic}
\end{algorithm}

\subsection{Definitions of Message Passing Layer and the UNet}

\label{appendix:mp_layer}

The spatial component of the model updates latent bubble embeddings through message passing over the bubble graph. For an edge \((i,j)\in\mathcal{E}\) at time \(t\), the message from bubble \(j\) to bubble \(i\) is computed as
\begin{equation}
    \mathbf{m}_{ij}^{t}
    =
    \varphi_{\mathbf{m}}
    \left(
    \mathbf{h}_i^{t},
    \mathbf{h}_j^{t},
    \mathbf{e}_{ij}^{t},
    \|\mathbf{x}_i^{t}-\mathbf{x}_j^{t}\|^2,
    \gamma_\tau(\tau)
    \right),
\end{equation}
where \(\mathbf{h}_i^{t}\) and \(\mathbf{h}_j^{t}\) are latent node embeddings, \(\mathbf{e}_{ij}^{t}\) denotes edge features, \(\mathbf{x}_i^{t}-\mathbf{x}_j^{t}\) is the relative displacement, and \(\gamma_\tau(\tau)\) is the embedding of the flow-matching time.

Messages are aggregated over neighboring bubbles,
\begin{equation}
    \bar{\mathbf{m}}_i^{t}
    =
    \sum_{j\in\mathcal{N}(i)}
    \mathbf{m}_{ij}^{t},
\end{equation}
and the hidden state is updated with a residual MLP block,
\begin{equation}
    \mathbf{h}_i^{t\,\prime}
    =
    \mathrm{LayerNorm}
    \left(
    \mathbf{h}_i^{t}
    +
    \varphi_{\mathbf{h}}
    \left(
    \mathbf{h}_i^{t},
    \bar{\mathbf{m}}_i^{t},
    \gamma_\tau(\tau)
    \right)
    \right).
\end{equation}
Here, \(\varphi_{\mathbf{m}}\) and \(\varphi_{\mathbf{h}}\) are MLPs with SiLU activations. The layer updates latent node features using geometric pairwise information; The layer preserves translational equivariance by constructing edge features from pairwise relative displacements and distances rather than absolute coordinates.

Our temporal convolution layer follows the U-Net design of \cite{brinke2026, dhariwal2021diffusionmodelsbeatgans} with the same parameters. The U-Net takes the latent node embeddings h together with a 16-dimensional sinusoidal embedding of the flow-matching time $\tau$. Since the trajectory data also depends on physical time within the window, we add a sinusoidal frame-index embedding to $\mathbf{h}$. The U-Net applies convolutions along the temporal axis over the T frames of each trajectory. Its output is an updated latent representation $\mathbf{h}$, from which a two-layer $\varphi_{\mathbf{v}}$ predicts the velocity-field update $\Delta \mathbf{v}$.

\subsection{Hyperparameters} \label{appendix:hyperparam}
The hyperparameters used to train and evaluate the model are included in Table~\ref{appendix:hyper}. We use AdamW with default parameters and a ReduceLROnPlateau scheduler based on the minimum validation loss, with a decay factor of 0.5 and patience of 60 epochs. The table lists the graph connectivity, trajectory window length, observed prefix length, and the prior-noise parameter s, which controls the additional noise in the informed prior by scaling the velocity variance estimated from the observed prefix.

\begin{table}[h]
    \centering
    \caption{Training hyperparameters for the experiments.}
    \begin{scriptsize}
    \setlength{\tabcolsep}{3pt}
    \begin{tabular}{lccccccccccc}
    \toprule
    \textbf{Model} 
    & \textbf{SH order} 
    & \textbf{connectivity} 
    & \textbf{layers} 
    & \(\boldsymbol{\tau}\) \textbf{dist.} 
    & \(\boldsymbol{s}\) 
    & \textbf{lr} 
    & \textbf{epochs} 
    & \textbf{val/test} 
    & \textbf{batch} 
    & \textbf{hidden dim} 
    & \textbf{window/prefix} \\
    \midrule
    Model \((L=5)\) 
    & 5 
    & 16 mm 
    & MP + U-Net 
    & $\sqrt{\mathcal{U}[0,1]}$ 
    & 2 
    & 5e-4 
    & 400 
    & 0.05/0.15 
    & 4 
    & 64 
    & 60/20 ms \\

    \textit{w/o} GNN 
    & 5 
    & -- 
    & U-Net 
    & $\sqrt{\mathcal{U}[0,1]}$ 
    & 2 
    & 5e-4 
    & 400 
    & 0.05/0.15 
    & 4 
    & 64 
    & 60/20 ms \\

    Model \((L=3)\) 
    & 3 
    & 16 mm 
    & MP + U-Net 
    & $\sqrt{\mathcal{U}[0,1]}$
    & 2 
    & 5e-4 
    & 400 
    & 0.05/0.15 
    & 4 
    & 64 
    & 60/20 ms \\
    \bottomrule
    \end{tabular}
    \end{scriptsize}
    \label{appendix:hyper}
\end{table}

\subsection{Compute resources} \label{appendix:res}
\textbf{Data Generation.} Simulations were run as single CPU processes on Debian Trixie or Ubuntu 24.04 desktop/workstations with 24-core AMD CPUs. Runtime varied substantially with the gas volume fraction and simulation stability. Lower gas fractions, such as \(\varepsilon\)=5\% or 10\%, typically required more than 30 days to generate over 1.5 s of simulated time, while higher gas fractions generally required at least 15 days for comparable physical durations.

\textbf{Model training and experiments.} All model experiments were run on a single NVIDIA Tesla V100 GPU. The prior baseline required less than 30 minutes to run, while the other model configurations required approximately 10-13 hours of training depending on the spherical harmonic order and whether message passing was used. Inference was comparatively inexpensive and took only a few minutes per experiment.



\subsection{Data Generation Pipeline}
Multiple simulations were run concurrently on any system alongside other activities. A daemon was created to detect the simulator writing an output file containing the Eulerian fields and original bubble meshes, and would then trigger the conversion program \cite{sph_bubble_convert}. The conversion of the bubble shapes to spherical harmonics was done ad hoc on the same machines, in parallel (up to 48 threads using OpenMP in C++, or multiprocessingPool in Python), after which the resulting coefficients were stored and the original output file was removed to save space. 

\newpage
\subsection{Visualizations} \label{app:viz}

\begin{figure}[h!]
\vspace{-3em}
    \centering
    \includegraphics[width=1\linewidth]{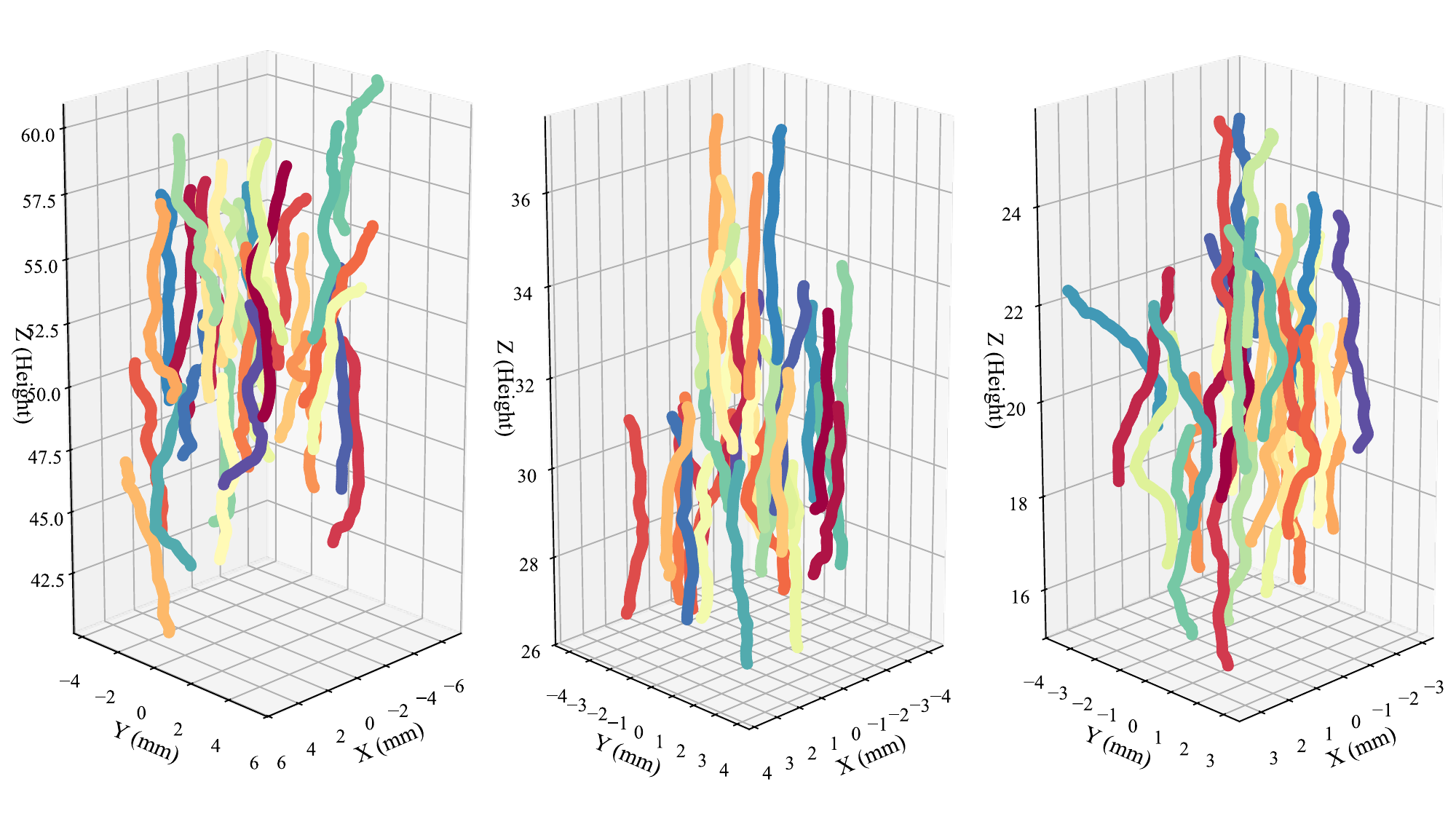}
    \caption{Visualizations of centered 32 bubble centroid trajectories. Duration is 200 ms, number of timesteps is 100, filtered to exclude periodic boundary jumps.}
    \label{fig:centroid_viz}
\end{figure}

\vspace{-3em}
\begin{figure}[h!]
    \centering
    \includegraphics[width=0.9\linewidth]{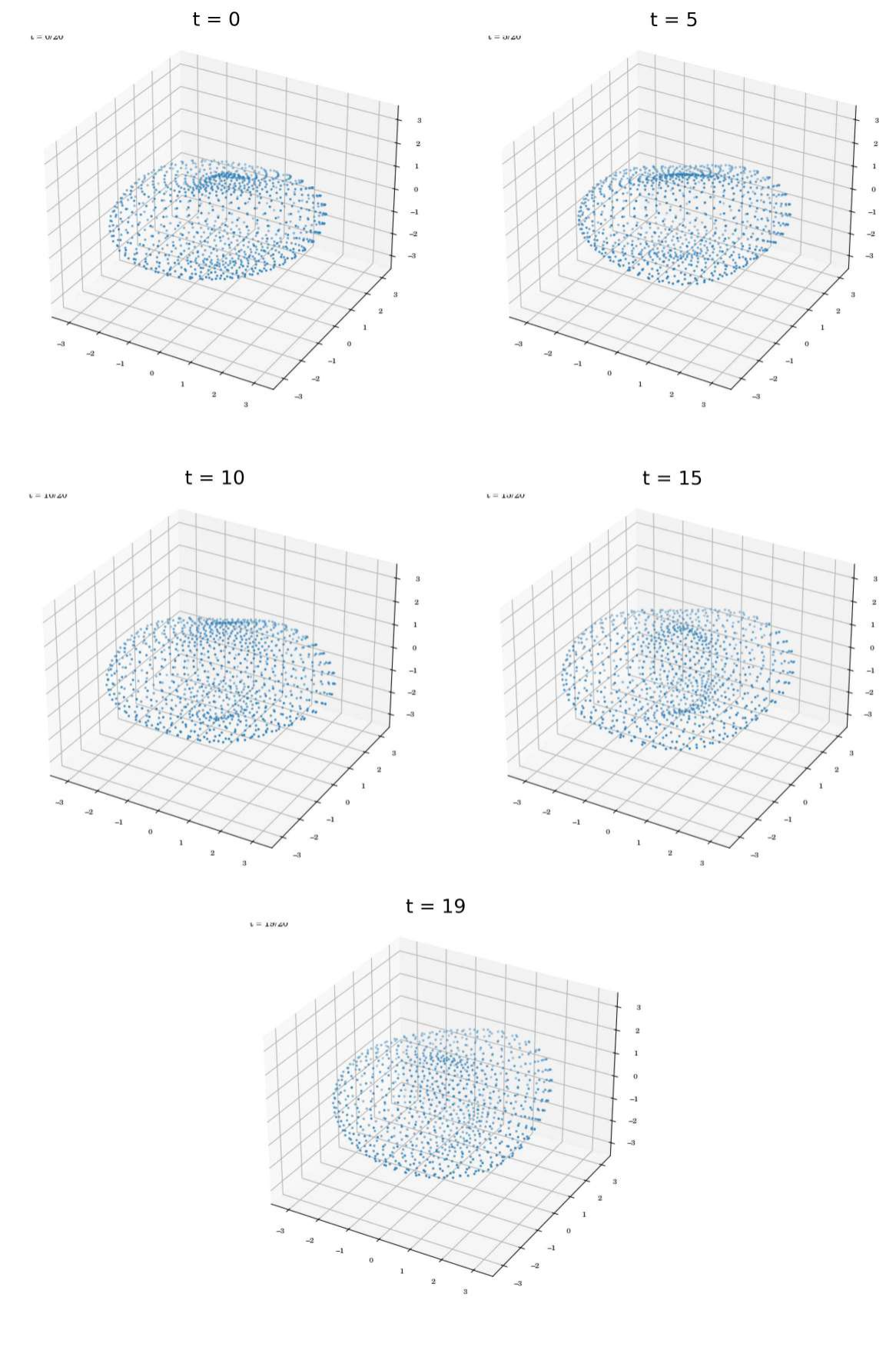}
    \caption{Visualizations of the shape of one bubble changing over time.}
    \label{fig:points}
\end{figure}


\newpage
\newpage

\end{document}